\documentclass[10pt,twocolumn,letterpaper]{article}

\usepackage{cvpr}              
\usepackage{graphicx}
\usepackage{multirow}
\usepackage{algpseudocode} 
\usepackage{amsmath} 
\usepackage{amssymb}
\usepackage{wrapfig}
\usepackage{graphicx}
\usepackage{caption}
\usepackage{subcaption}
\usepackage{floatflt}
\usepackage{booktabs} 
\usepackage{array} 
\usepackage{xcolor}
\usepackage{lipsum}
\usepackage{colortbl}
\usepackage{algorithm}
\usepackage{amssymb}
\usepackage{float} 
\usepackage{afterpage}
\usepackage{stfloats}
\usepackage{multicol}
\usepackage{pifont}
\usepackage[T1]{fontenc}

\definecolor{cvprblue}{rgb}{0.21,0.49,0.74}
\usepackage[pagebackref,breaklinks,colorlinks,allcolors=cvprblue]{hyperref}

\def\paperID{1737} 
\def\confName{CVPR}
\def\confYear{2025}

\title{UniAvatar: Taming Lifelike Audio-Driven Talking Head Generation with Comprehensive Motion and Lighting Control}

\author{
Wenzhang Sun\textsuperscript{1*}, 
Xiang Li\textsuperscript{2*}, 
Donglin Di\textsuperscript{1},
Zhuding Liang\textsuperscript{1},
Qiyuan Zhang\textsuperscript{3},\\
Hao Li\textsuperscript{1}
Wei Chen\textsuperscript{1}, 
Jianxun Cui\textsuperscript{1},
\\
\textsuperscript{1}Li Auto, 
\textsuperscript{2}Harbin Institute of Technology, 
\textsuperscript{3}Zhejiang University,
\\
}

\begin{document}


\maketitle

\renewcommand{\thefootnote}{\fnsymbol{footnote}}
\footnotetext[1]{Equal contribution.}




\begin{abstract}
Recently, animating portrait images using audio input is a popular task. Creating lifelike talking head videos requires flexible and natural movements, including facial and head dynamics, camera motion, realistic light and shadow effects. Existing methods struggle to offer comprehensive, multifaceted control over these aspects. In this work, we introduce UniAvatar, a designed method that provides extensive control over a wide range of motion and illumination conditions. Specifically, we use the FLAME model to render all motion information onto a single image, maintaining the integrity of 3D motion details while enabling fine-grained, pixel-level control. Beyond motion, this approach also allows for comprehensive global illumination control. We design independent modules to manage both 3D motion and illumination, permitting separate and combined control. Extensive experiments demonstrate that our method outperforms others in both broad-range motion control and lighting control. Additionally, to enhance the diversity of motion and environmental contexts in current datasets, we collect and plan to publicly release two datasets, DH-FaceDrasMvVid-100 and DH-FaceReliVid-200, which capture significant head movements during speech and various lighting scenarios.

\end{abstract}

\section{Introduction}
\begin{figure}[h]  
    \centering
    \includegraphics[width=\linewidth]{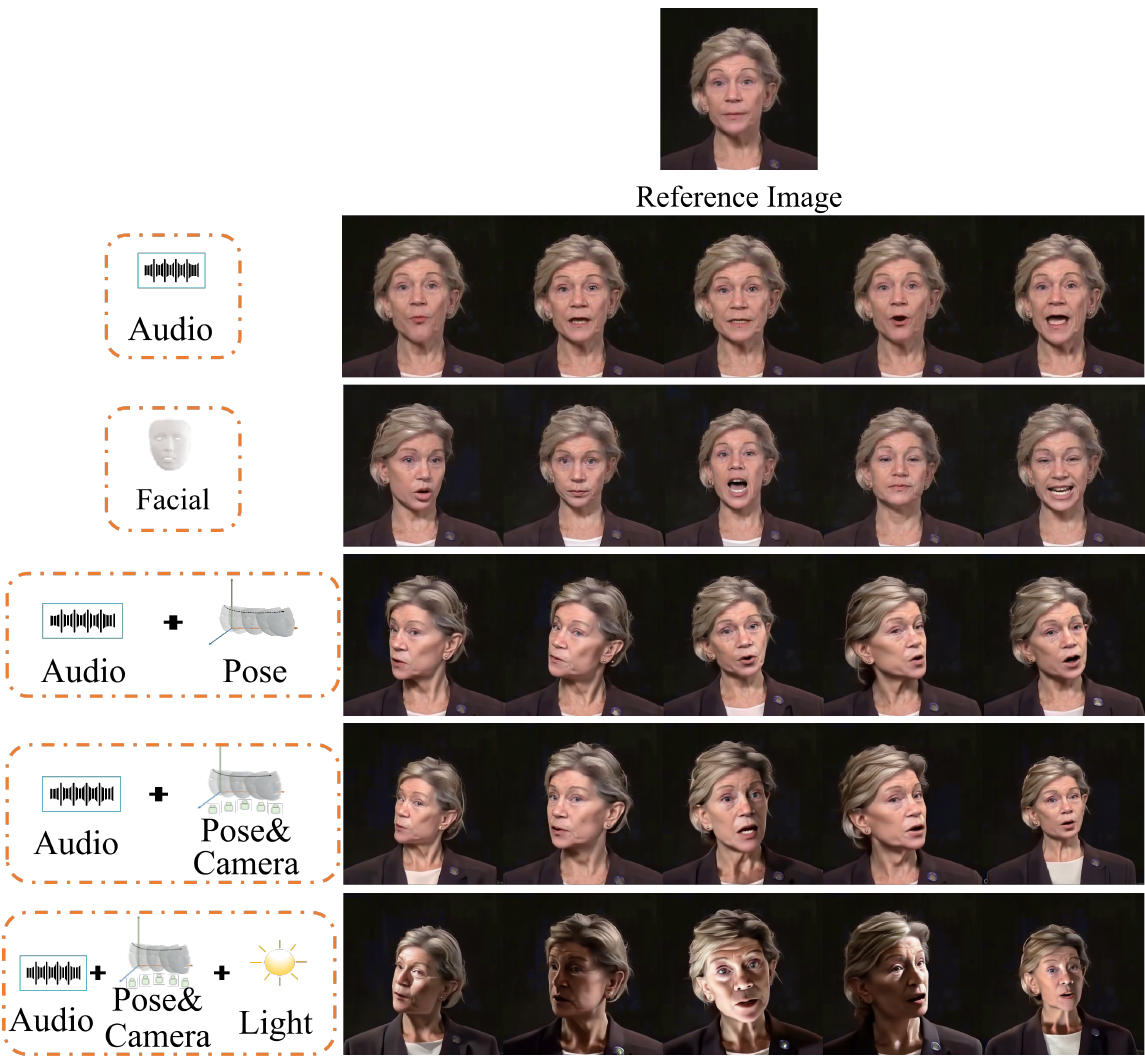}  
    \caption{Showcases under various control signals. Our method enabling different motion controls without failure during extensive movements, as well as allowing the flexible generation under different lighting conditions.}
    \label{fig:showcase1}
    \vspace{-5mm}

\end{figure}

\begin{figure*}[!ht]  
    \centering
    \includegraphics[width=\linewidth]{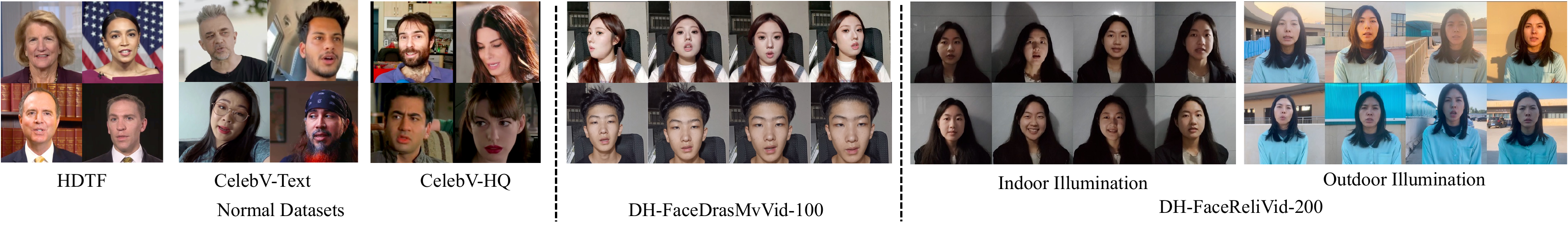}  
    \vspace{-5mm}
    \caption{Showcase of publicly available datasets and our proposed datasets: We refer to datasets like HDTF and DH-FaceVid-1k as normal datasets, which contain a wide range of identity information. In contrast, our datasets offer more extensive motion variations under the same identity (DH-FaceDrasMvVid-100) and more diverse lighting conditions under the same identity (DH-FaceReliVid-200).}
    \label{fig:dataset}
    \vspace{-5mm}

\end{figure*}

Diffusion models have made significant strides in the field of generative models, demonstrating exceptional proficiency in generating high-fidelity images \cite{dhariwal2021diffusion,ho2020denoising,blattmann2023stable,rombach2022high,sohl2015deep}. These advances have had a profound impact on video generation, enabling exploration of their use to create dynamic and engaging visual narratives \cite{bar2024lumiere,ho2022video,zhang2023i2vgen}. A key area of focus is human-centric video generation, particularly portrait video synthesis \cite{ma2023dreamtalk,stypulkowski2024diffused,sun2023vividtalk,zhang2023sadtalker} and human animation \cite{guo2023animatediff,hu2024animate}, due to their potential applications in the creation and production of digital avatars. Among these, 'talking head' video generation has garnered particular interest, with the aim of producing head videos synchronized with input audio, capturing intricate facial expressions, head movements, and precise lip synchronization.

Creating lifelike talking head videos requires not only accurate lip synchronization but also flexible and natural motion. However, translating audio into motion is challenging because of the ambiguous one-to-many mapping. Current methods often separate motions by using different representations as conditions to control the generation process. For example, EchoMimic \cite{Echomimic} combines facial landmarks with audio to control head pose, providing more control than audio alone. V-Express \cite{V-Express} notes that variations in the strength of the control signal can interfere with each other, so it simplifies the control by focusing on three keypoints to manage the pose of the head. MegActor$-\sum$ \cite{MegActor} uses a driving video to guide head motion during generation. Building on this, VASA1 \cite{xu2024vasa} integrates multiple motion control signals within a DiT architecture, enabling comprehensive control over generated motions such as gaze, head pose, and camera movement. However, such extensive control often requires additional conditioning signals, complicating the network structure and increasing training complexity. Moreover, relying on two-dimensional control conditions limits the model’s potential, making it challenging to model environmental interactions and causing instability in talking head video generation with significant motion.

This paper introduces UniAvatar, a novel method for talking head generation that offers comprehensive control over various types of motion while interacting with environmental lighting, enabling generation under diverse lighting conditions (Figure \ref{fig:showcase1}). To achieve more flexible control over both motion and illumination conditions, we use two distinct signals to control the generation process. Specifically, for controlling various motion signals, unlike VASA1 \cite{xu2024vasa}, which independently handles different motion signals for different conditions, UniAvatar employs Motion-aware Rendering to integrate multiple control signals into a single guidance image. Furthermore, we utilize a 3D Motion Module to extract motion information and inject it into the attention module of the denoising model, enabling pixel-wise control over motion. Additionally, to achieve global lighting control, we use Illumination-aware Rendering to generate global Illumination condition and enhance the impact of lighting on the output through a Masked-Cross-Source Sampling Strategy, thus enabling comprehensive control over global illumination.


To overcome the limitations of existing datasets and achieve comprehensive control over motion and lighting, we have curated two spcific talking head datasets: DH-FaceDrasMvVid-100 and DH-FaceReliVid-200. DH-FaceDrasMvVid-100 includes 100 hours of speaking videos featuring extensive movements, both in-plane and out-of-plane. DH-FaceReliVid-200 comprises 200 hours of video, capturing individuals under diverse indoor and outdoor lighting conditions, with each subject recorded in 8 different lighting environments. Our key contributions can be summarized as follows:

\begin{itemize}
    \item We propose UniAvatar, a novel framework capable of generating characters with flexible motion and illumination. To the best of our knowledge, this is the first portrait animation method that simultaneously controls different types of motion (head, camera movement and facial motion) and global illumination.
    \item Using FLAME model \cite{flame}, we integrate multiple motion representations into a single image and employ a 3D Motion module for pixel-wise control, enabling stable generation over large-scale movements.
    \item By using illumination-aware rendering and a masked sampling strategy, we achieve flexible illumination control over the generated videos.
    \item To address the limitations of existing datasets lacking motion and lighting diversity, we specifically collected two datasets: DF-FaceDrasMvVid-100 and DH-FaceReliVid-200, which will be released for open-source research.
\end{itemize}

\section{Related Work}

\textbf{Diffusion Models.} Diffusion-based generative models have recently become pivotal in computational creativity due to their versatility in multimedia tasks \cite{sohl2015deep,song2020score,song2020denoising}. These models excel in synthesizing new images, enhancing existing visuals, producing dynamic video content, and creating intricate 3D digital constructs \cite{zhang2023adding,dong2023open,liu2023zero,sargent2023zeronvs,shi2023mvdream,long2024wonder3d,qiu2024richdreamer,fu2025geowizard,dai2023animateanything,harvey2022flexible,ho2022video,khachatryan2023text2video,yang2023diffusion,zhu2024champ}. A prime example is the SD model \cite{blattmann2023stable}, which uses a UNet framework to generate images based on textual descriptions, trained on large-scale datasets linking text to images. The pre-trained diffusion models offer flexibility across static and dynamic visual media. Innovative approaches are enhancing these models by combining UNet with Transformer-based designs, like in DiT, to improve text-conditioned video generation. Furthermore, diffusion models are gaining popularity for generating realistic animated portraits, known as 'talking heads'.


\noindent \textbf{Talking Head Generation.} Significant progress has been made in generating lifelike talking head videos \cite{guo2021ad,ma2023styletalk,sun2021speech2talking,wang2023lipformer,wang2021audio2head,wang2022one,yu2023talking,zhang2021flow,zhou2020makelttalk}. EchoMimic \cite{Echomimic} introduces additional keypoint control beyond audio, allowing for more flexible facial generation. V-Express \cite{V-Express} observes that excessive control signals can interfere with generation quality, so it uses only three keypoints to control head movement. VASA-1 \cite{xu2024vasa} constructs separate motion latents for different control signals, enabling comprehensive control over head movement, facial expressions, and camera motion. Building on this, our work adds environmental lighting control alongside motion control, achieving both independent and coupled control of multiple signals through carefully designed control mechanisms.


\noindent \textbf{Portrait Relighting.} In face relighting, a common approach is to decompose an image into intrinsic components—such as lighting, albedo, and surface normals—and then recompose it with modified lighting \cite{barron2014shape,le2019illumination,sengupta2018sfsnet,shu2017neural,tewari2021monocular,wang2008face,sun2023neural,papantoniou2023relightify,qiu2024relitalk}. Diffusion-based methods consider lighting as a global control condition, enabling image generation under specific illumination. DiFaReli \cite{difareli}, for instance, uses DDIM's near-perfect inversion to achieve portrait re-illumination, while IC-Light \cite{iclight} generates characters and backgrounds in various lighting settings based on control inputs like text and images. Unlike these image-level relighting methods, our approach produces dynamic talking head videos with stable backgrounds under specified ambient lighting conditions.

\section{Data Collection and Filteration}
\begin{table}[t]
\LARGE
\setlength\tabcolsep{4pt}
\centering
\caption{Summary of our proposed dataset and filtered data}
\label{Table:dataset}
\resizebox{\linewidth}{!}{
\begin{tabular}{lccccc}
\toprule
Datasets &
  \begin{tabular}[c]{@{}c@{}}Filtered \\ IDs\end{tabular} &
  \begin{tabular}[c]{@{}c@{}}Filtered \\ Hours\end{tabular} &
  \begin{tabular}[c]{@{}c@{}}Identity \\ Diversity\end{tabular} &
  \begin{tabular}[c]{@{}c@{}}Motion \\ Diversity\end{tabular} &
  \begin{tabular}[c]{@{}c@{}}Light \\ Information\end{tabular} \\ \midrule
HDTF \cite{zhang2021flow}&
  362 &
  16 &
  \ding{72}\ding{72} &
  \ding{72} &
  - \\
CelebV-HQ \cite{zhu2022celebv}& 
5000 &
9.3 &
  \ding{72}\ding{72}\ding{72} &
  \ding{72} &
  - \\
 CelebV-Text \cite{yu2023celebv}& 
2500 &
11.5 &
  \ding{72}\ding{72}\ding{72} &
  \ding{72} &
  \ding{72} \\ 
DH-FaceDrasMvVid-100 &
  50 &
  100 &
  \ding{72} &
  \ding{72}\ding{72}\ding{72} &
  - \\
DH-FaceReliVid-200 &
  200 &
  200 &
  \ding{72}\ding{72} &
  \ding{72} &
  \ding{72}\ding{72}\ding{72} \\ \bottomrule
\end{tabular}
}
\vspace{-5mm}
\end{table}
The talking head generation model intrinsically possesses the capacity to effectively scale with large datasets. Nonetheless, its efficacy is contingent upon the availability of high-quality data. Current datasets have notable limitations: (1) Limited Head Movement: Head motions are typically restricted to about 20 degrees, offering minimal 3D information and leading to failures in large-scale movement generation. (2) Uniform Lighting: Lighting conditions lack diversity within the same identity, limiting the model's ability to adapt to environmental lighting variations. More diverse movement and lighting data are needed to improve the model’s real-world generalization.

To meet the demand for more realistic and versatile talking head generation and to address current dataset limitations, we developed two novel datasets: \textbf{DH-FaceDrasMvVid-100} and \textbf{DH-FaceReliVid-200} (Figure \ref{fig:dataset}). DH-FaceDrasMvVid-100 includes 100 hours of talking videos featuring extensive head movements, categorized into in-plane movements (lateral and pitch rotations) and out-of-plane movements (distance variations between the speaker and camera). DH-FaceReliVid-200 provides 200 hours of talking videos under varied lighting, with recordings of the same subjects in eight distinct indoor and eight outdoor lighting conditions. These datasets enable training or fine-tuning of models to improve performance. We combined open-source data with our proprietary collection for training (Table \ref{Table:dataset}). Detailed dataset metrics and descriptions can be found in the supplementary materials.

\section{Method}

\begin{figure*}[t]  
    \centering
    \includegraphics[width=\linewidth]{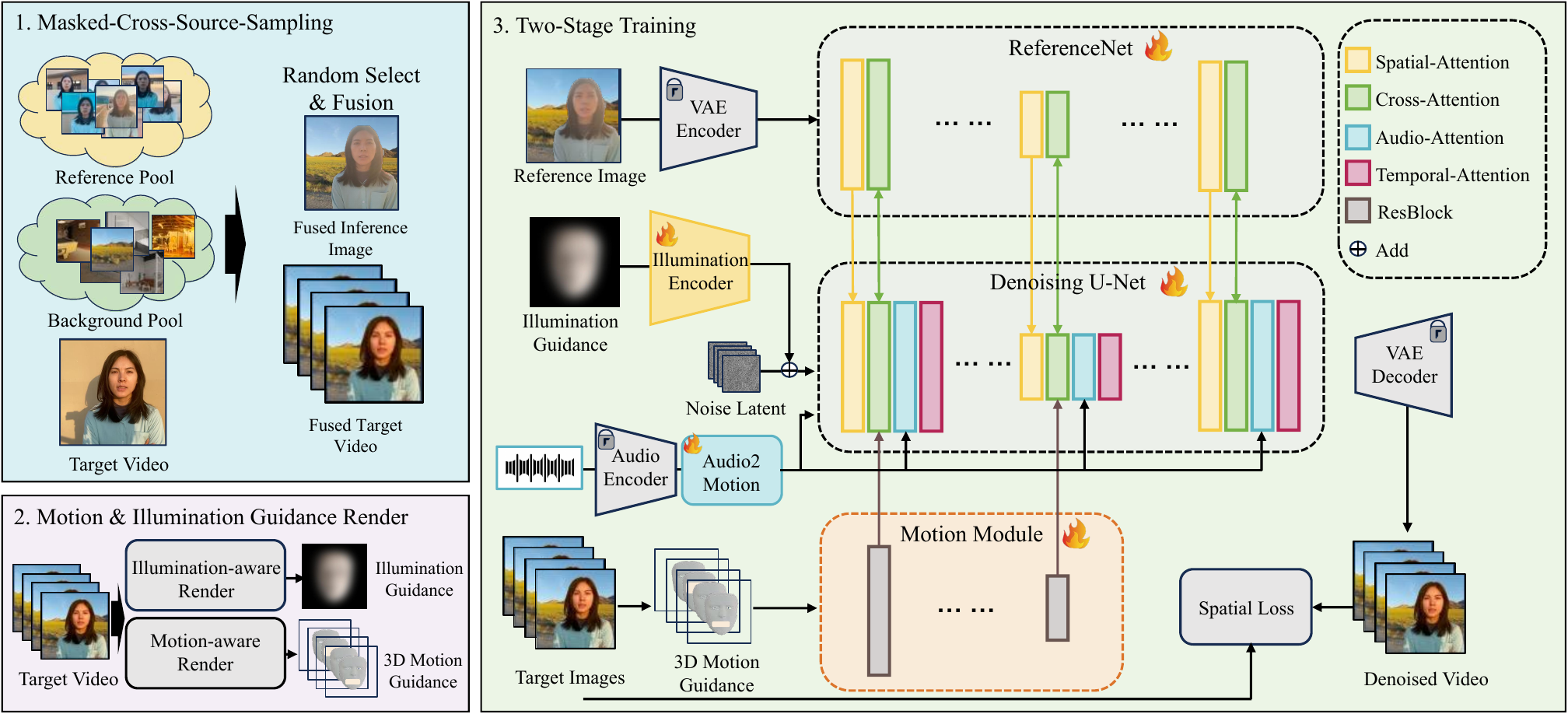}  
    \caption{The overall frame work of UniAvatar. We use a Masked-Cross-Source Sampling strategy to learn the lighting information and ensure background stability. To enable independent and combined control over different conditions, we utilize separate render for each condition and dedicated modules for injecting motion and illumination conditions.}
    \label{fig:model}
    \vspace{-5mm}
\end{figure*}

\begin{figure}[t]  
    \centering
    \includegraphics[width=\linewidth]{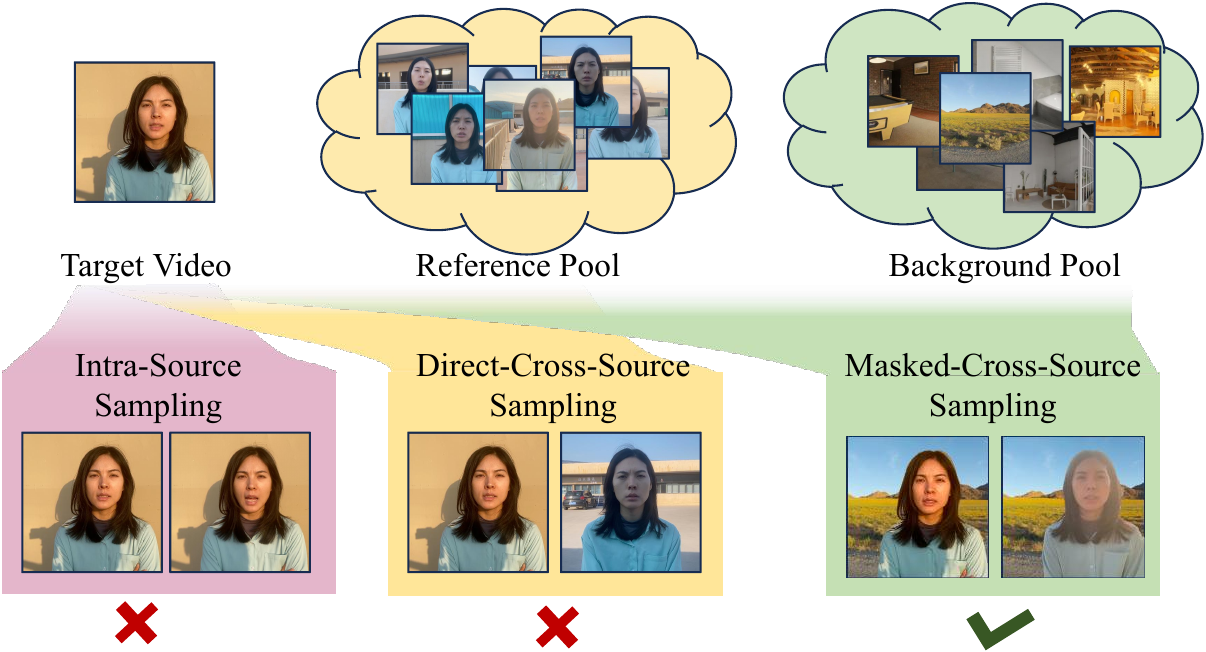}  
    \caption{Visualization of the sampling strategy. We sample from different source videos under the same identity. To ensure background stability, we build a database of 500 background images and randomly select and composite new images.}
    \label{fig:masked-sample_strategy}
    \vspace{-5mm}
\end{figure}

\subsection{Model Overview}

Given one portrait image $x$ and a sequence of speech clip $s = [s_{1}, \dots, s_{N}]$, the model aims to generate a talking video clip $[x_{1}, \dots, x_{N}]$. Furthermore, given head motion $m$, camera motion $c$ and Illumination $l$, the model can further refine the generated video to align with these specified conditions. In Section 4.2, we introduce the overall network architecture. In Section 4.3, we discuss how to decouple different control signals to achieve individual or combined control over generation. In Section 4.4, we introduce the detailed training and inference strategy.

\subsection{Framework}

Figure \ref{fig:model} illustrates the architecture of the proposed approach. The ReferenceNet module encodes the reference image, capturing the visual appearance of both the portrait and the associated background. To maintain background consistency under varying lighting conditions, we implemented a Masked-Cross-Source Sampling strategy. The input audio embedding is derived from a 12-layer wav2vec network \cite{Schneider_Baevski_Collobert_Auli_2019}. Additionally, we developed an illumination encoder and a motion module independently to extract motion information. The illumination encoder is a lightweight convolutional neural network, while the motion module is based on the framework of \cite{dhariwal2021diffusion}, utilizing a modified U-Net architecture built with residual block stacks interspersed with self-attention layers. By synthesizing these diverse conditioning elements within the denoising U-Net \cite{blattmann2023stable}, the model generates frames that uphold visual coherence with the reference image while flexibly representing varied and subtle motions and illumination effects. Temporal alignment is achieved through multiple self-attention blocks, each optimized to handle the sequential elements across video frames.
\subsection{Guidance Disentanglement}
We use the FLAME face model \cite{flame} to separately construct 3D motion and illumination guidance. The 3D motion guidance represents different motion information using a single image, while the illumination guidance contains only global lighting information. This rendered image contains rich 3D information and allows for pixel-level control over the generated result.  We use an off-the-shelf single-view 3D face reconstruction method, DECA \cite{deca}. Given a face image, DECA predicts the 3D face shape, camera pose and spherical harmonic lighting (SH) coefficients.

\noindent\textbf{3D Motion Guidance.} The 3D motion encoder is dedicated to extracting features related to head, facial, and camera motion. During rendering, the motion-aware renderer sets the lighting spherical harmonics coefficients to a fixed value. The rendering formula is as follows:
\begin{equation}
\label{cp}
P_{i,j}^{M} =A\odot \sum_{k=1}^{9}I^{'}_{k}H_{K} \mathcal{N} (\beta^{'}, \theta^{'}, \psi^{'}, c^{'})
\end{equation}
Where $\beta^{'}, \theta^{'}$ and $\psi ^{'}$ is the facial pose, shape and expressions of the target image; $c^{'} \in \mathbb{R}^{1+2}$ represents orthographic camera parameters; $I^{'}_{k} \in \mathbb{R}^{3}$ is manually set spherical harmonic lighting parameters designed to ensure consistent lighting across all 3D motion guidance;and the SH basis is defined as $H_{k} : \mathbb{R}^{3} \to \mathbb{R}$ ; $P^{M} \in \mathbb{R}^{H \times W \times 3}$ is rendered 3D motion condition. Detailed descriptions are provided in the supplementary material. After encoding, the image is integrated with features that have undergone spatial attention through a U-Net encoder. The integration mechanism can be expressed as: 
\begin{equation}
s_{i}^{'} = s_{i} \odot tanh(p_{i})
\end{equation}
We denote the output of the residual module or attention module prior to each downsampling convolution in the UNet network as $p_{i} $, and the output of each spatial attention layer within the denoising UNet as $s_{i}$. In the denoising UNet, we substitute the output $s_{i}$ of each spatial attention layer with $s_{i}^{'}$, and subsequently input $s_{i}^{'}$ into the  cross-attention layers. This approach enables pixel-wise control, allowing for improved motion generation and maintaining stability even during wide-range movements.

\noindent\textbf{Illumination Guidance.} The illumination guidance encodes the lighting into a guidance image that is independent of motion and contains only lighting information. During the training process, We extract the illumination parameter $I_{k} \in \mathbb{R}^{3}$ of the target video, use the camera parameters $c \in \mathbb{R}^{1+2}$, pose $\beta$, shape $\theta$, and expression $\psi$ under zero-pose. The rendering formula is as follows:
\vspace{-3 pt}
\begin{equation}
P_{i,j}^{I} =A\odot \sum_{k=1}^{9}I_{k}H_{K} \mathcal{N} (\beta, \theta, \psi, c)
\end{equation}
\vspace{-3 pt}
To reduce the influence of illumination conditions on motion, we apply a Gaussian blur to $P^{I} \in \mathbb{R}^{H\times W\times 3}$ . The illumination encoder a lightweight network. Initially, we extract features of the illumination conditions through a set of convolution layers. Subsequently, an attention module is applied after the convolution layers to accurately capture lighting-related semantic information. To preserve the integrity of the pretrained denoising U-Net model, we use convolution layers initialized to zero as the output layer to extract features of the lighting conditions. These lighting features are then combined with noise latent representations before being input into the denoising fusion module.


\noindent\textbf{Audio-to-Motion.} To enhance the richness of facial movements, we incorporate an adaptive layer normalization mechanism into the denoising U-Net architecture. We extract the facial expression coefficients $\beta$. After aligned with the audio features, we random select one of the embeddings and using a zero-initialized multilayer perceptron (MLP) to produce scaling ($\gamma$) and shifting ($\epsilon$) parameters.The adaptive layer normalization is applied between the cross-attention layer and the audio attention layer within the denoising U-Net.

\begin{figure*}[t]  
    \centering
    \includegraphics[width=\linewidth]{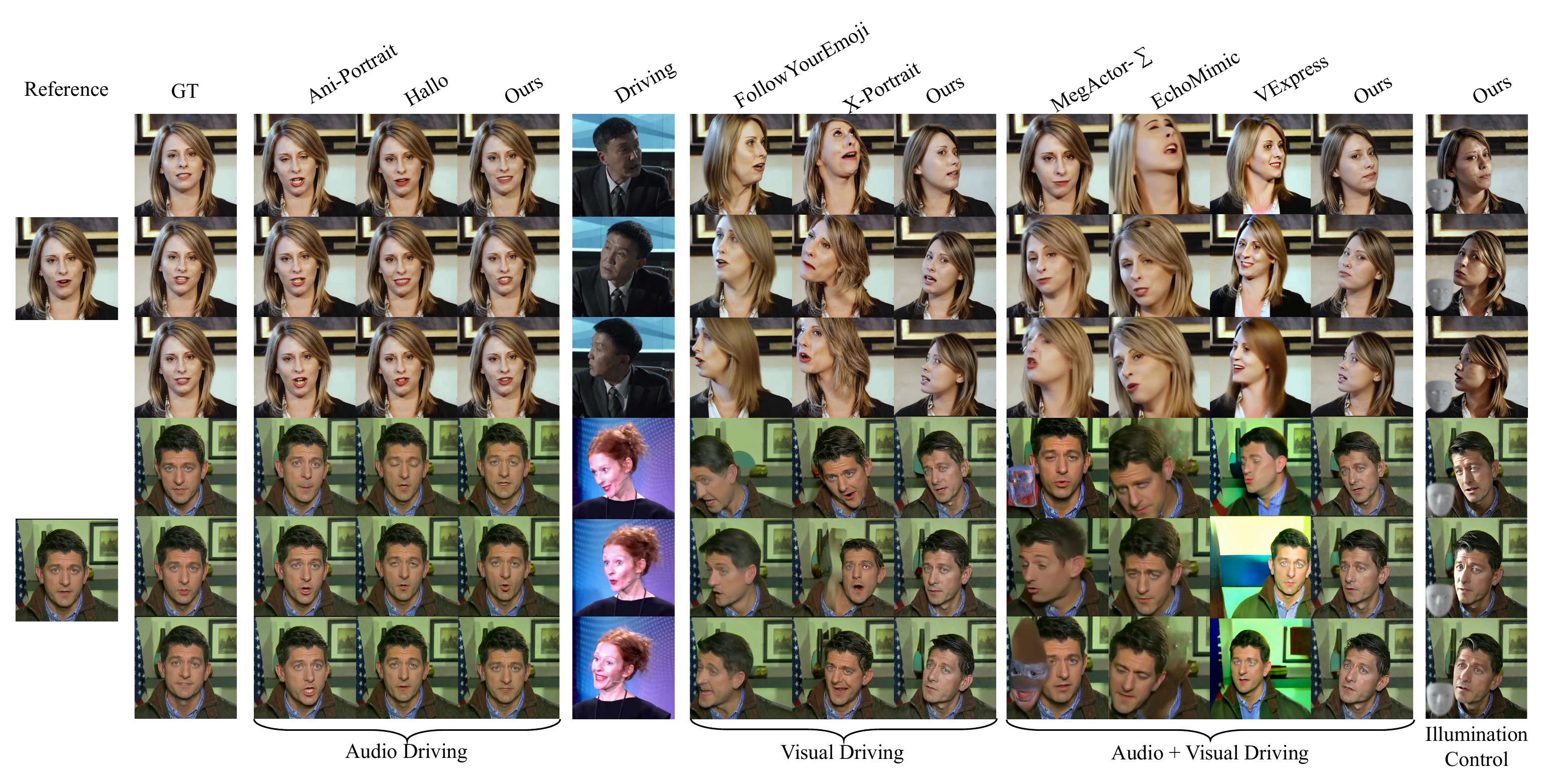}  
    \vspace{-7mm}
    \caption{Visual comparisons with different methods. Results demonstrate that UniAvatar surpasses other method across multiple control modalities. UniAvatar maintains stability even during large movements and provides flexible global illumination control.}
    \label{fig:main}
    \vspace{-5mm}

\end{figure*}

\subsection{Training and Inference}


\noindent\textbf{Masked-Cross-Source Sampling(MCSS).} We conceptualize illumination as a global control, focusing on generating videos under specified lighting conditions based on a reference image. Given a reference image and global illumination, it is challenging to ensure lighting continuity in the video without three-dimensional information. We employed a masked cross-source sampling strategy (Figure \ref{fig:masked-sample_strategy}) to amplify the effect of illumination and further applied image-level constraints to enhance continuity. Specifically, for each ID, we constructed a candidate pool to select the reference image and target video. Additionally, due to the significant background variation in outdoor lighting, this can lead to inconsistencies in the generated background and inaccuracies in illumination. We collected 500 background images, randomly selected from them, and used a facial segmentation algorithm to blend the foreground and background. This approach allows us to create data pairs for each ID that share the same background but differ in illumination.

\noindent\textbf{Conditions Mask and Dropout.} We integrate multiple control conditions—3D motion, illumination, and audio—to guide our model. To disentangle the overlapping information among these conditions, we employ tailored masking and dropout strategies during training. Specifically, for the 3D motion condition, we mask the lips region with a probability of 0.4 and apply dropout at a rate of 0.2. To prevent conflicts between illumination and facial controls, we blur the illumination condition using a Gaussian convolution kernel with a radius of 15.

\noindent\textbf{Progressive Training.} We employ a two-stage training process to optimize the components of the framework. In Stage 1, the model generates motion video frames based on a reference image, driven audio, and 3D motion condition, keeping the VAE encoder/decoder and facial image encoder parameters fixed. Training focuses on the spatial and cross-attention modules within ReferenceNet and the denoising U-Net, with illumination guidance initialized as learnable parameters. We integrate with the illumination dataset after 10,000 steps to refine lighting information. In Stage 2, the illumination encoder, motion module, and attention modules in ReferenceNet and the denoising U-Net remain static, emphasizing video sequence generation. This stage primarily trains hierarchical audio-visual cross-attention to establish the connection between audio as motion guidance and visual elements like lip, expression, and pose. Additionally, pre-trained motion modules from AnimateDiff are introduced to improve temporal coherence, with a randomly selected frame from the video clip as the reference image.

\begin{figure}[t]  
\centering
\includegraphics[width=\linewidth]{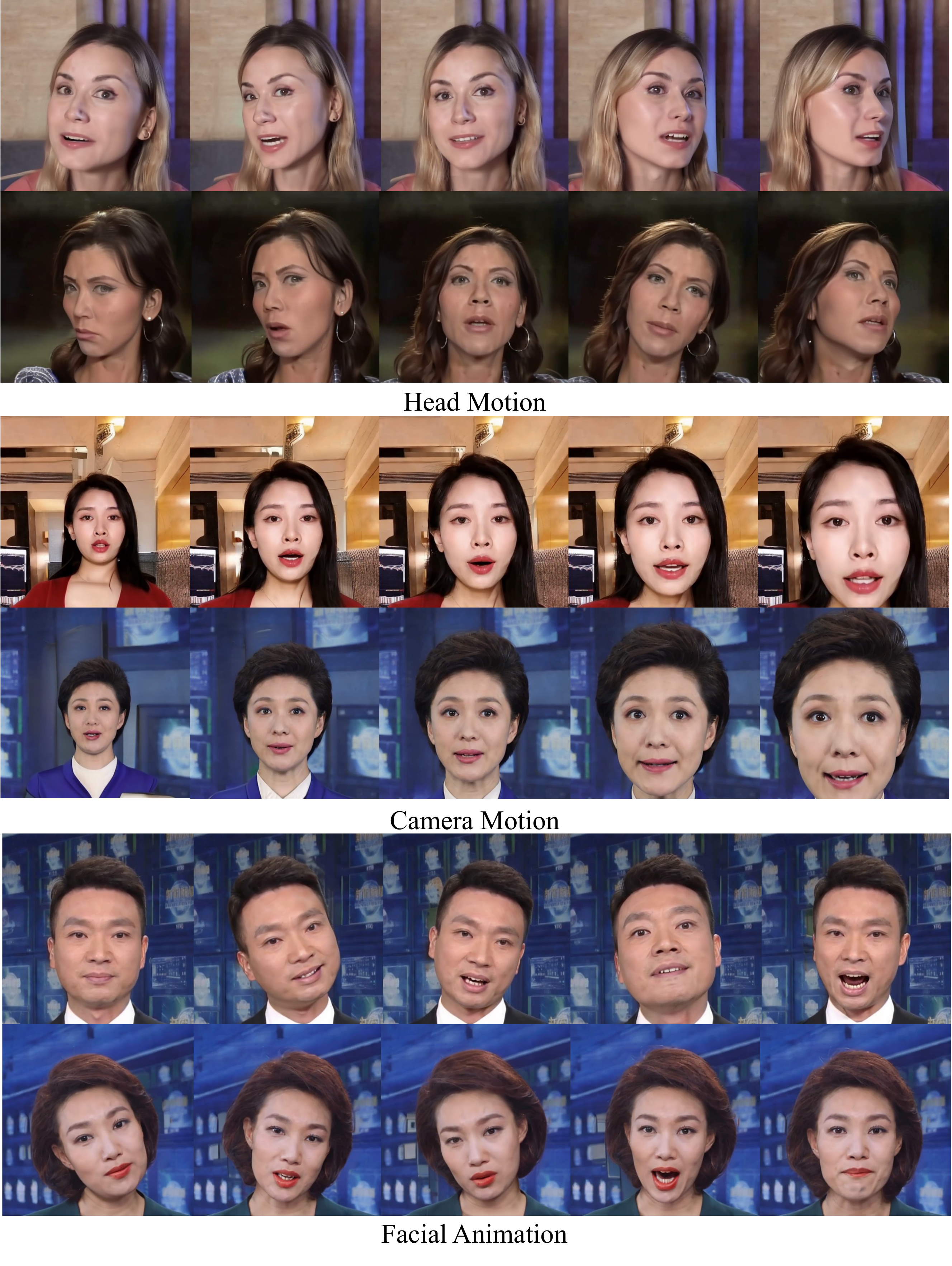}  
\vspace{-3mm}
\caption{Visualization of video generated under different motion conditions.
}
    \label{fig:motion}
    \vspace{-5mm}

\end{figure}

\noindent\textbf{Loss Defination.} To ensure continuity in the reconstruction results, we use LPIPS loss in addition to latent loss to capture similarity at the pixel level. The loss function $L_{obj}$ during the training process is summarized as:
\begin{equation}
    L_{obj} =L_{latent}+\lambda L{spatial}
\end{equation}
Where $L_{latent}$ is the objective function guiding the denoising process. With timestep $t$ is uniformly sampled from $\left \{ 1, ..., T  \right \} $. The objective is to minimize the error between the true noise \(\epsilon\) and the model-predicted noise \(\epsilon_{\theta}(z_{t}, t, c)\) based on the given timestep \(t\), the noisy latent variable \(z_t\), and the conditional information \(c\). The loss function during the training process is summarized as:
\begin{equation}
\mathcal{L} =\mathbb{E}_{t,c,z_{t},\epsilon}[||\epsilon-\epsilon_{\theta}(z_{t},t,c)||^2]
\end{equation}
$L{spatial}$ denotes the perceptual loss (LPIPS loss) used to further refine the details. Moreover, since the model struggles to converge at large time steps $t$, a time-step-aware function to reduce the weight for large $t$. The detailed objective function is shown as follows:
\vspace{-3pt}
\begin{equation}
L{spatial} = cosine(t*\pi/2T)LPIPS(I_{p},I_{GT})
\end{equation}
\noindent\textbf{Inference.} During inference, UniAvatar supports various control signals and their combinations, categorized as follows: (1) Audio-control: generating video based on an audio signal; (2) Motion-control: generating videos with additional head, facial or camera movement; (3) Illumination-control: adjusting environmental lighting effects based on a provided illumination signal.

\section{Experiment}
\begin{figure}[!t]  
\centering
\includegraphics[width=\linewidth]{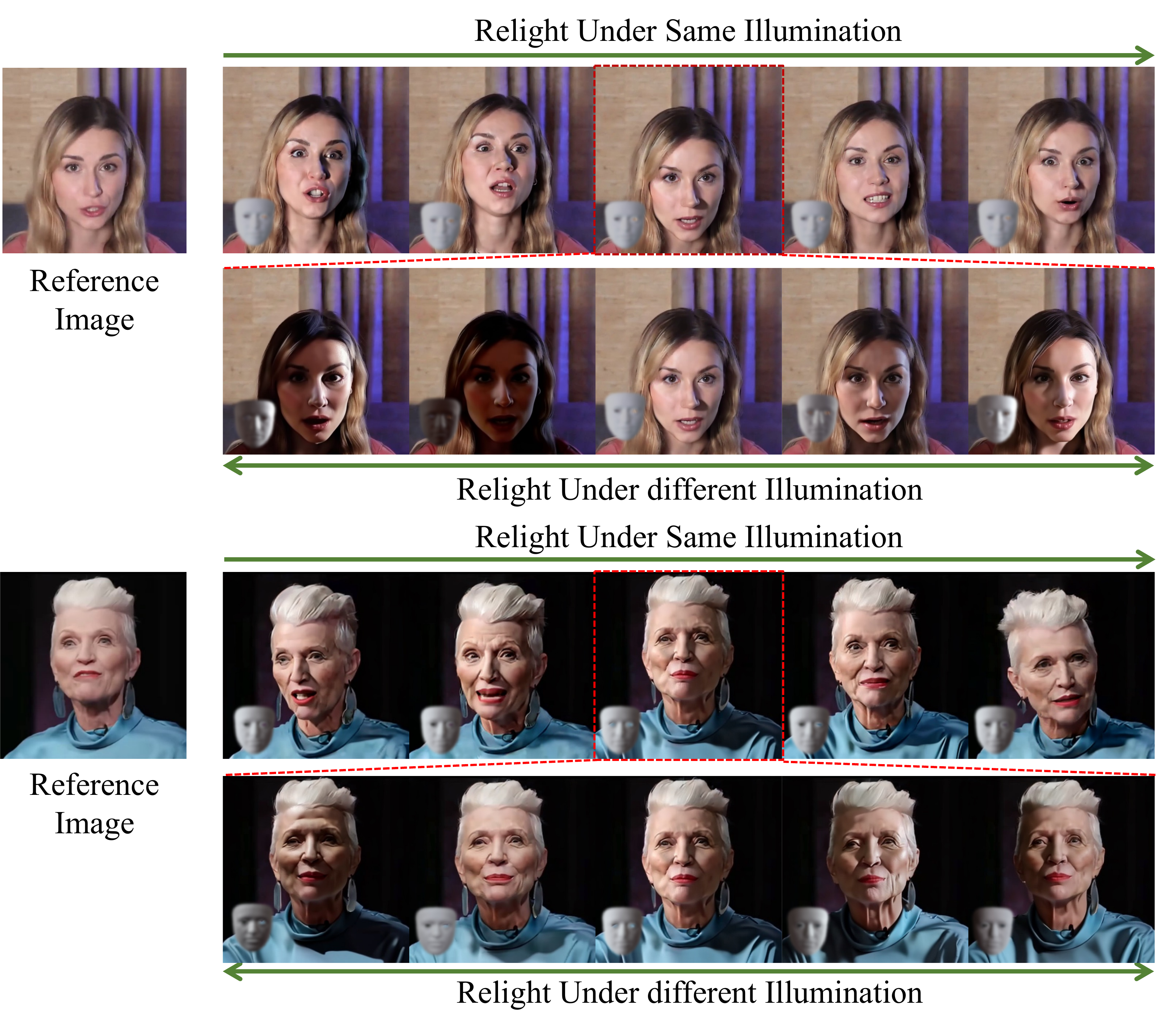}  
\caption{Video generation result under different illumination.}
    \label{fig:light}
    \vspace{-5mm}

\end{figure}

\subsection{Experiment setup}
\noindent\textbf{Datasets.} We utilizes the Filtered HDTF \cite{zhang2021flow}, CelebV-HQ \cite{zhu2022celebv} and CelebV-text \cite{yu2023celebv} dataset, along with DH-FaceDrasMvVid-100 and DH-FaceReliVid-200 for training. For testing, we isolate 20 subjects from both HDTF and DH-FaceDrasMvVid-100, randomly selecting 10 video clips per subject, each lasting 5-10 seconds. To evaluate the relighting results, 20 subjects are chosen from DH-FaceReliVid-200, with 10 videos per subject under 4 randomly selected lighting conditions out of 8, while the reference image is drawn from a different lighting condition with masked background. Further details are provided in the supplementary material.


\noindent\textbf{Evaluation Metrics.} We employ Peak Signal-to-Noise Ratio (PSNR), Learned Perceptual Image Patch Similarity (LPIPS), Fréchet Inception Distance (FID) \cite{FIDheusel2017gans}, Fréchet Video Distance (FVD) \cite{FVDunterthiner2018towards}, Average Keypoint Distance (AKD) \cite{AKDsiarohin2019animating,AKDsiarohin2019first}, and synchronization metrics Sync-C and Sync-D \cite{sync}. PSNR assesses image fidelity and structure, with higher scores indicating better quality, while LPIPS measures perceptual similarity, with lower scores being preferable. FID and FVD evaluate realism, where lower scores indicate greater similarity to real data. AKD measures the alignment of facial keypoints, with lower values indicating higher accuracy in reproducing facial expressions and movements. Sync-C and Sync-D \cite{sync} evaluate lip synchronization, where higher Sync-C and lower Sync-D scores reflect better audio-visual alignment. Since the dataset’s audio is in Chinese, Sync-C and Sync-D may not provide an entirely objective evaluation, so we used tailored combinations of evaluation metrics for each dataset.


\noindent\textbf{Baseline.} We compared our proposed method with publicly available implementations of AniPortrait \cite{wei2024aniportrait}, Hallo \cite{xu2024hallo}, MegActor$-\sum$ \cite{MegActor}, Follow-Your-Emoji \cite{ma2024follow}, X-Portrait \cite{xie2024x}, V-Express \cite{V-Express}, and EchoMimic \cite{Echomimic} on the HDTF, DH-FaceDrasMvVid-100, and DH-FaceReliVid-200 datasets. We also conducted a qualitative comparison to provide deeper insights into our method's performance and its ability to generate realistic, expressive talking head animations.



\subsection{Quantitative Results.}
We conducted a comprehensive comparison of existing diffusion-based methods driven by audio, video, and combined audio-video signals, highlighting UniAvatar's effectiveness in utilizing diverse control signals. To ensure rigorous evaluation across datasets, we selected appropriate metrics to showcase each method’s unique strengths.
\begin{figure}[!t]  
\centering
\includegraphics[width=\linewidth]{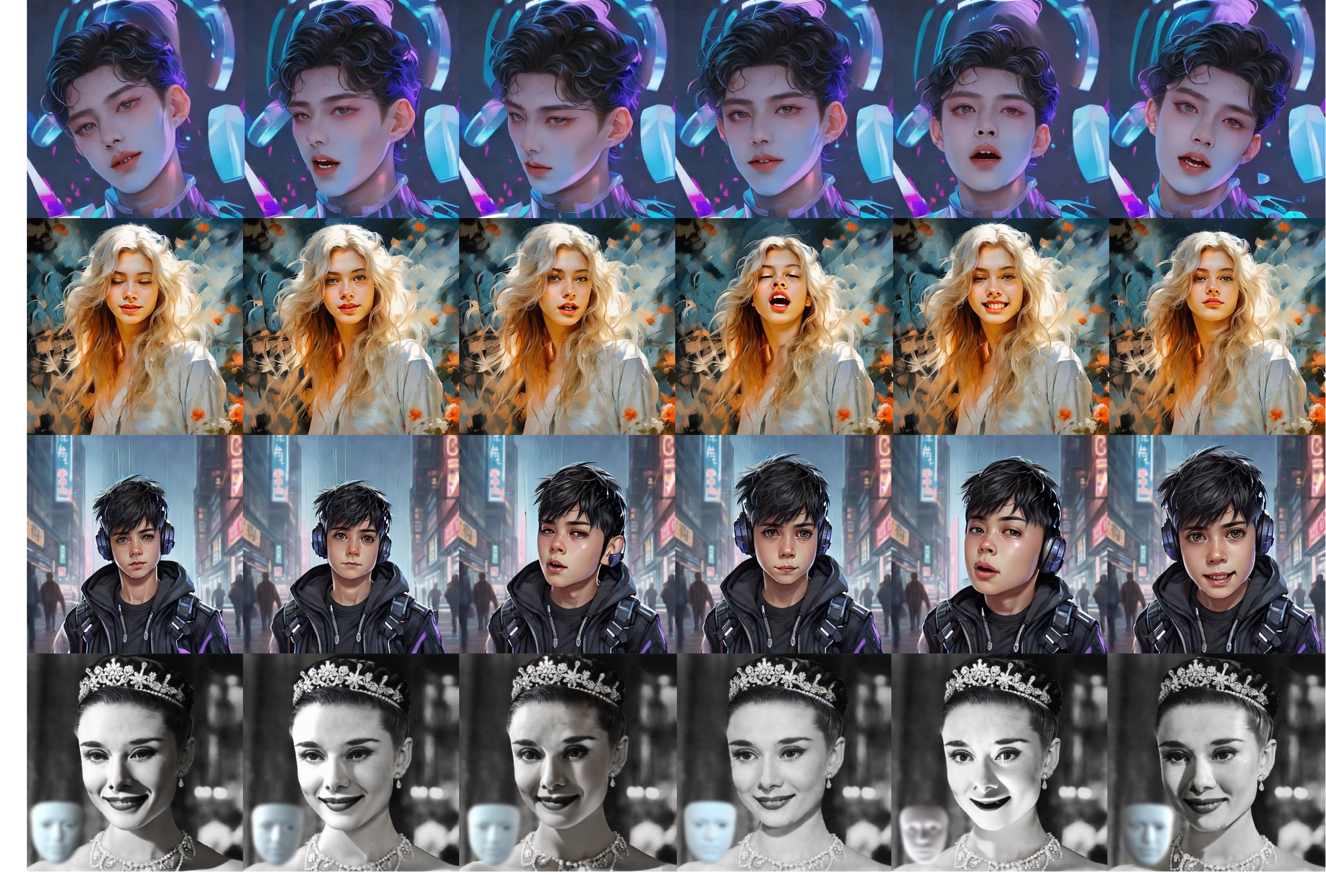}  
\caption{Video generation results of the proposed approach given different portrait styles.
}
    \label{fig:style}
    \vspace{-3mm}

\end{figure}

\begin{table}[t]
\centering
\caption{Comparison of various methods on the HDTF dataset.}\label{tab:hdtf}
\vspace{-3mm}
\resizebox{0.48\textwidth}{!}{
\begin{tabular}{clccccc}
\toprule
\multirow{2}{*}{\textbf{Modal}}                                            & \multirow{2}{*}{\textbf{Method}} & \multicolumn{5}{c}{\textbf{HDTF}} \\ \cline{3-7} 
                       &                   & \textbf{FID$\downarrow$}    & \textbf{FVD$\downarrow$}     & \textbf{LPIPS$\downarrow$} & \textbf{Sync-C$\uparrow$} & \textbf{Sync-D$\downarrow$} \\ \midrule
\multirow{3}{*}{Audio} & AniPortrait \cite{wei2024aniportrait}       & 36.826          & 476.818          & 0.219          & 5.977               & 9.899               \\
                       & Hallo \cite{xu2024hallo}             & \textbf{28.605} & 343.023          & 0.167          & \textbf{6.273}               & \textbf{8.735}               \\
                       & Ours          & 28.732          & \textbf{320.326} & \textbf{0.157} & 6.132               & 9.044               \\ \midrule
\multirow{3}{*}{Video} & Follow-Your-Emoji \cite{ma2024follow} & 29.111          & 296.195          & \textbf{0.107} & 6.297               & 9.832               \\
                       & X-Portrait \cite{xie2024x}        & 31.596          & 277.446          & 0.161          & 5.831               & 9.158               \\
                       & Ours          & \textbf{27.34}  & \textbf{274.573} & 0.113          & \textbf{6.547}               & \textbf{9.017}               \\ \midrule
\multirow{4}{*}{\begin{tabular}[c]{@{}c@{}}Audio\\ +\\ Video\end{tabular}} & V-Express\cite{V-Express}                        & 27.156  & 274.99  & 0.114 & 6.721 & 9.192 \\
                       & Echomimic \cite{Echomimic}         & 58.323          & 526.125          & 0.333          & \textbf{6.773}                & 9.257                 \\
                       & MegActor$-\sum$ \cite{MegActor}          & 56.116          & 421.395          & 0.346          & 5.973               & \textbf{8.321}               \\
                       & Ours   & \textbf{27.013} & \textbf{268.341} & \textbf{0.103} & 6.657               & 8.976               \\ \bottomrule
\end{tabular}
}
\vspace{-3mm}

\end{table}

\noindent\textbf{Comparison on the HDTF dataset.} Table \ref{tab:hdtf} indicates that UniAvatar outperforms other methods across multiple metrics under various control conditions. The Sync-C and Sync-D metrics show a slight decline due to the Chinese talking head dataset we used.
\begin{table}[t]
\centering
\caption{Comparison of various methods on the DH-FaceDrasMvVid-100 datasets.}\label{tab:dras}
\vspace{-3mm}
\resizebox{0.48\textwidth}{!}{
\begin{tabular}{clccccc}
\toprule
\multirow{2}{*}{\textbf{Modal}} &
  \multirow{2}{*}{\textbf{Method}} &
  \multicolumn{5}{c}{\textbf{DH-FaceDrasMvVid-100}} \\ \cline{3-7} 
 &
   &
  \textbf{FID$\downarrow$} &
  \textbf{FVD$\downarrow$} &
  \textbf{LPIPS$\downarrow$} &
  \textbf{PSNR$\uparrow$} &
  \textbf{AKD$\downarrow$} \\ \midrule
\multirow{3}{*}{Audio} &
  AniPortrait \cite{wei2024aniportrait} &
  \textbf{57.297} &
  856.851 &
  0.273 &
  16.708 &
  30.44 \\
 &
  Hallo \cite{xu2024hallo} &
  59.733 &
  922.327 &
  \textbf{0.258} &
  \textbf{17.755} &
  \textbf{27.819} \\
 &
  Ours &
  58.273 &
  \textbf{831.273} &
  0.271 &
  17.231 &
  27.835 \\ \midrule
\multirow{3}{*}{Video} &
  Follow-Your-Emoji \cite{ma2024follow} &
  40.784 &
  567.516 &
  0.153 &
  21.985 &
  3.176 \\
 &
  X-Portrait \cite{xie2024x} &
  68.762 &
  536.386 &
  0.275 &
  16.893 &
  25.46 \\
 &
  Ours &
  \textbf{37.187} &
  \textbf{332.531} &
  \textbf{0.147} &
  \textbf{22.103} &
  \textbf{2.673} \\ \midrule
\multirow{4}{*}{\begin{tabular}[c]{@{}c@{}}Audio\\ +\\ Video\end{tabular}} &
  V-Express \cite{V-Express} &
  48.398 &
  322.461 &
  0.183 &
  21.76 &
  4.984 \\
 &
  Echomimic \cite{Echomimic} &
  74.126 &
  758.585 &
  0.351 &
  16.585 &
  5.109 \\
 &
  MegActor$-\sum$ \cite{MegActor} &
  75.69 &
  689.922 &
  0.471 &
  13.334 &
  26.816 \\
 &
  Ours &
  \textbf{36.644} &
  \textbf{311.491} &
  \textbf{0.122} &
  \textbf{24.477} &
  \textbf{2.459} \\ \bottomrule
\end{tabular}
}
\vspace{-3mm}

\end{table}

\noindent\textbf{Comparison on the DH-FaceDrasMvVid-100 dataset.} DH-FaceDrasMvVid-100 contains a large set of facial videos with extensive motion, enabling UniAvatar to generate more stable results under conditions of wide-range movement. As shown in Table \ref{tab:dras}, UniAvatar significantly outperforms all other methods across all metrics.

\begin{table}[t]
\centering
\caption{Comparison of various methods on the DH-FaceReliVid-200 dataset.}\label{tab:reli}
\vspace{-3mm}
\resizebox{0.48\textwidth}{!}{
\begin{tabular}{clcccc}
\toprule
\multirow{2}{*}{\textbf{Modal}} &
  \multirow{2}{*}{\textbf{Method}} &
  \multicolumn{4}{c}{\textbf{DH-FaceReliVid-200}} \\ \cline{3-6} 
                       &                   & \textbf{FID$\downarrow$} & \textbf{FVD$\downarrow$} & \textbf{LPIPS$\downarrow$} & \textbf{PSNR$\uparrow$} \\ \midrule
\multirow{3}{*}{Audio} & AniPortrait \cite{wei2024aniportrait}        & 117.428      & 1218.315     & 0.395          & 14.211        \\
                       & Hallo \cite{xu2024hallo}             & 113.394      & 991.949      & 0.386          & 14.311        \\
                       & Ours          & 115.538      & 1131.725     & 0.391          & 14.435        \\ \midrule
\multirow{3}{*}{Video} & Follow-Your-Emoji \cite{ma2024follow} & 113.493      & 1193.268     & 0.383          & 14.392        \\
                       & X-Portrait \cite{xie2024x}        & 119.712      & 1045.806     & 0.462          & 13.323        \\
                       & Ours          & 115.372      & 1032.654     & 0.361          & 15.451        \\ \midrule
\multirow{4}{*}{\begin{tabular}[c]{@{}c@{}}Audio\\ +\\ Video\end{tabular}} &
  V-Express \cite{V-Express} &
  109.858 &
  1049.287 &
  0.258 &
  16.031 \\
                       & Echomimic \cite{Echomimic}         & 135.702      & 1443.772     & 0.417          & 14.212        \\
                       & MegActor$- \sum$ \cite{MegActor}          & 126.267      & 1105.748     & 0.462          & 13.31         \\
                       & Ours        & 105.027      & 973.276      & 0.273          & 16.701        \\ \hline
Audio+Video+Light &
  Ours &
  \textbf{74.372} &
  \textbf{607.725} &
  \textbf{0.175} &
  \textbf{22.521} \\ \bottomrule
\end{tabular}
}
\vspace{-3mm}

\end{table}

\noindent\textbf{Comparison on the DH-FaceReliVid-200 dataset.} The comparison in Table \ref{tab:reli} focuses on the lighting effects in the generated videos, with the reference image selected across different videos and the background masked out. Unlike other methods, which cannot generate videos with varied lighting conditions, UniAvatar outperforms all others across all metrics.

\subsection{Qualitative Results}
\noindent\textbf{Visual Comparisons.} Figure \ref{fig:main} presents a visual comparison of UniAvatar against other methods across different control modalities, demonstrating superior performance under various conditions. Notably, UniAvatar provides more stable and accurate results during wide-range motion. Furthermore, when given global lighting conditions, UniAvatar can generate talking head videos with specific lighting effects—a capability not achievable by other methods.


\noindent\textbf{Visualization on Motion Control.} Figure \ref{fig:motion} illustrates UniAvatar's visual control over various motion conditions, showcasing its superior performance under wide-range head, camera, and facial movements. The results demonstrate that UniAvatar maintains stability in the generated output across diverse, large-scale movements.

\noindent\textbf{Visualization on Illumination Control.} Figure \ref{fig:light} demonstrates UniAvatar's relight capabilities, presenting both continuous video results under specified ambient lighting and results across different lighting conditions. The findings indicate that UniAvatar can flexibly adapt to various ambient lighting settings, with illumination effects that extend beyond the face to include well-lit areas in the neck and chest.



\noindent\textbf{Animation of Different Portrait Styles.} Figure \ref{fig:style} illustrates the ability of our method to handle diverse input types, such as oil paintings, anime images and portraits generated by other models. These results underscore the versatility and effectiveness of our approach in adapting to various artistic styles.


\subsection{Ablation Study}

\begin{figure}[!t]  
\centering
\includegraphics[width=0.9\linewidth]{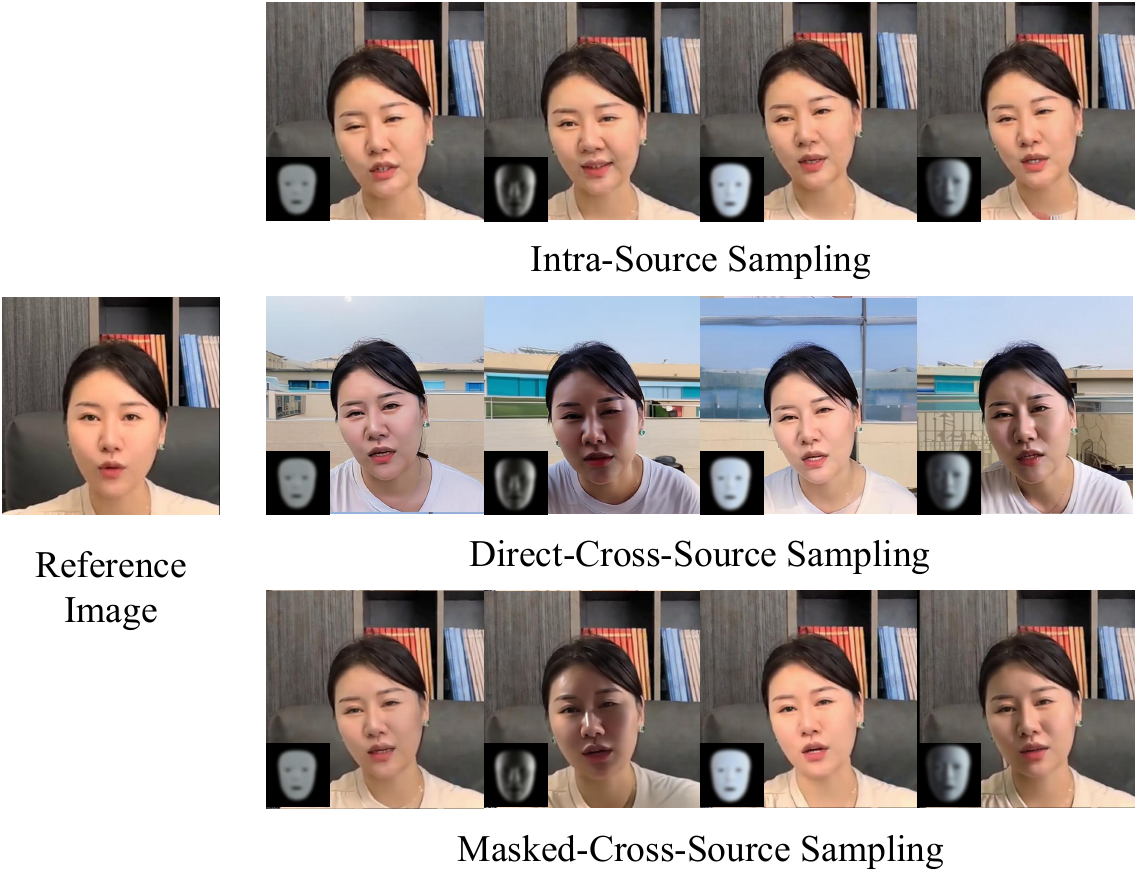}  
\vspace{-3mm}
\caption{Visualization of different sampling strategy. Results demonstrate that the proposed sampling strategy can ensure background stability while relighting.
}
    \label{fig:samplestra}
    \vspace{-3mm}

\end{figure}

\noindent\textbf{Masked-Cross-Source Sampling.} We introduce the MCSS strategy to enhance illumination controllability and background stability. We compared three sampling methods: intra-source, direct-cross-source, and masked-cross-source. The experimental results demonstrate that our proposed approach effectively manages illumination and maintains background stability (Figure \ref{fig:samplestra}).

\begin{figure}[!t]  
\centering
\includegraphics[width=\linewidth]{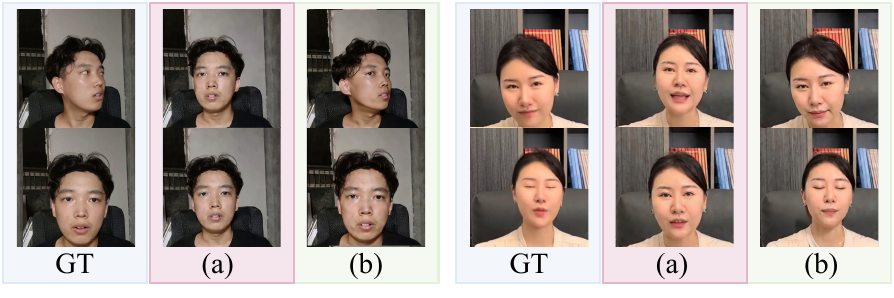}  
\caption{Visualization of different injection methods. We compared (a) injection at noise locations; (b) injection at spatial-attention locations. Injection at spatial-attention locations provides better control effects during motion.
}
    \label{fig:inject}
    \vspace{-5mm}

\end{figure}

\noindent\textbf{Motion Injection Method.} We compared two different motion control methods: (1) adding latent motion directly at the noise level (referencing EchoMimic \cite{Echomimic}); (2) extracting features using a motion module and integrating them after spatial attention. Figure \ref{fig:inject} indicates that the method used in this paper enables pixel-level control, allowing for a wider range of motion control.

\begin{figure}[!t]  
\centering
\includegraphics[width=\linewidth]{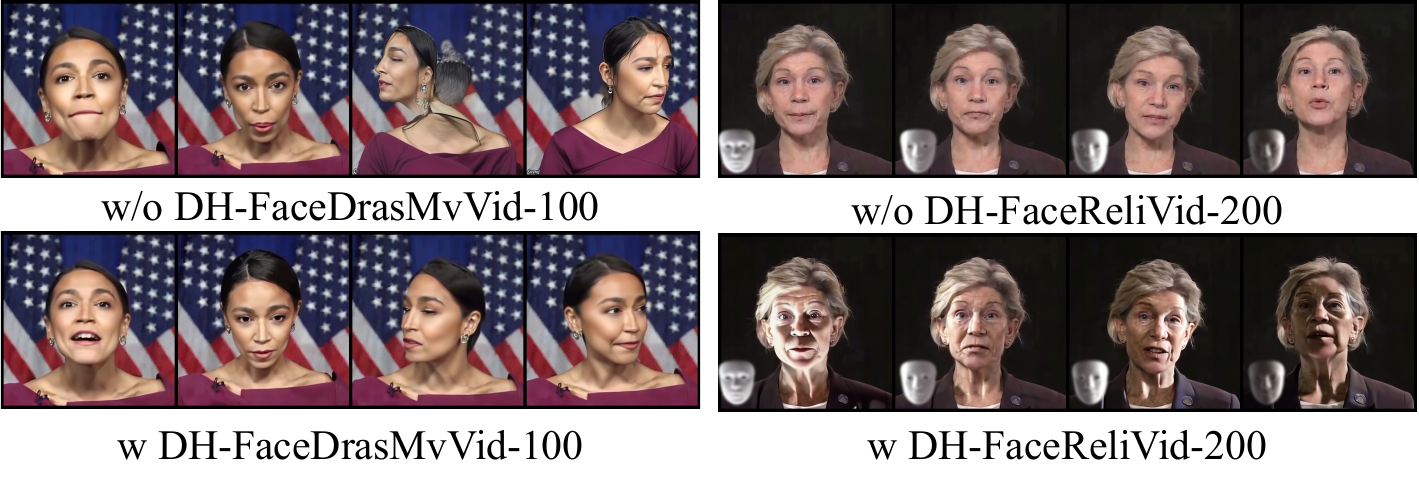}  
\vspace{-4mm}
\caption{Visualization of dataset efficiency: Training with the proposed dataset enhances stability in wide-range motion and improves illumination effects.
}
    \label{fig:dataeff}
    \vspace{-5mm}

\end{figure}

\noindent\textbf{Dataset Efficientcy.} To evaluate the effectiveness of the data set, we performed qualitative comparison experiments (Figure \ref{fig:dataeff}). The results verified that DH-FaceDrasMvVid-100 significantly enhances the stability of generated results during large movements, while DH-FaceReliVid-200 allows for controllable lighting in generated videos.

For more in-depth analysis, please refer to the supplementary material.


\section{Limitations}

Our method still has certain limitations in the following areas: (1) Under certain strong lighting conditions, white edges may appear due to inaccuracies in the segmentation method, and remnants of the original background may persist in images during the MCSS process. (2) UniAvatar encounters challenges in generating facial adornments. For instance, when a character wearing glasses moves extensively, the results are not yet satisfactory. Our future efforts will aim to solve these problems.
\section{Conclusion}

This paper introduces UniAvatar, a novel end-to-end approach for generating lifelike talking head videos. UniAvatar provides extensive control over a wide range of motion and illumination conditions, exceeding the capabilities of existing methods. By utilizing 3D priors from the FLAME model, UniAvatar facilitates precise control of diverse motion signals and flexible adjustments to environmental lighting effects. To improve the motion and lighting diversity of existing talking head datasets, we have curated two new datasets: DH-FaceDrasMvVid-100 and DH-FaceReliVid-200, which will be made publicly available for research purposes.
 \small \bibliographystyle{ieeenat_fullname} \bibliography{main}


\end{document}


\maketitle
\renewcommand\thesection{\Alph{section}}
\renewcommand{\thefigure}{A.\arabic{figure}}
\setcounter{figure}{0}
\renewcommand{\thetable}{A.\arabic{table}}
\setcounter{table}{0}

\section{Overview}
In this Appendix, we present:
\begin{itemize}

\item Section \ref{sec:B} :Implementation details
\item Section \ref{sec:C} :Network architectures
\item Section \ref{sec:D} :Additional Results

\end{itemize}

\section{Implementation details}\label{sec:B}
\begin{table*}[]
\LARGE
\setlength\tabcolsep{4pt}
\centering
\caption{Detailed information about the collected datasets}
\label{Table:dataset}
\resizebox{\linewidth}{!}{
\begin{tabular}{ccccccccccccc}
\toprule
\multirow{2}{*}{\begin{tabular}[c]{@{}c@{}} Datasets\end{tabular}} & \multicolumn{4}{c}{Meta Information} & \multirow{2}{*}{\begin{tabular}[c]{@{}c@{}}Large-scale \\ Head Move\end{tabular}} & \multirow{2}{*}{\begin{tabular}[c]{@{}c@{}}Distance \\ Change\end{tabular}} & \multirow{2}{*}{Action} & \multirow{2}{*}{Emotion} & \multirow{2}{*}{\begin{tabular}[c]{@{}c@{}}HDRI \\ Map\end{tabular}} & \multirow{2}{*}{\begin{tabular}[c]{@{}c@{}}FLAME \\ Model\end{tabular}} & \multirow{2}{*}{Environment} & \multirow{2}{*}{Format} \\ \cline{2-5}
                                                                             & Ids.   & Reso.      & FPS.   & Dura  &                                                                                   &                                                                             &                         &                          &                                                                      &                                                                         &                              &                         \\ \hline
DH-FaceDrasMvVid-100                                                         & 50    & 512*512+   & 30     & 100 h   & \ding{51}                                                                                 & \ding{51}                                                                           & \ding{51}                       & \ding{51}                        & \ding{55}                                                                    & \ding{51}                                                                       & Lab                          & Video                   \\
DH-FaceReliVid-200                                                           & 200    & 512*512+   & 30     & 200 h   & \ding{55}                                                                                 & \ding{55}                                                                           & \ding{51}                       & \ding{51}                        & \ding{51}                                                                    & \ding{51}                                                                       & Wild/Lab                     & Video                   \\ \toprule
\end{tabular}
}
\label{tab:dataset}
\end{table*}
\subsection{Dataset}
We utilized the HDTF \cite{zhang2021flow} dataset and filtered 5000 and 2500 video data corresponding to IDs from CelebV-HQ \cite{zhu2022celebv}  and CelebV-Text \cite{yu2023celebv} to enhance ID information. The dataset proposed in this paper, DH-FaceDrasMvVid-100 and DH-FaceReliVid-200, provides a wide range of motion and lighting information. We provided detailed information about the datasets collected in this study in Table \ref{tab:dataset}. 

During testing, we conducted separate evaluations on the HDTF, DH-FaceDrasMvVid-100, and DH-FaceReliVid-200 datasets. On the HDTF dataset, we evaluated the FID \cite{FIDheusel2017gans}, FVD \cite{FVDunterthiner2018towards}, LPIPS, Sync-C, and Sync-D \cite{sync} metrics, focusing primarily on the overall quality of reconstructed images and videos, as well as lip-sync accuracy. Since the Sync-C and Sync-D metrics cannot provide accurate results for Chinese audio, alternative metrics were specifically chosen for the DH-FaceDrasMvVid-100 and DH-FaceReliVid-200 datasets. Given that the DH-FaceDrasMvVid-100 dataset includes extensive human motion, we selected PSNR and AKD \cite{AKDsiarohin2019animating,AKDsiarohin2019first} metrics to evaluate the accuracy of reconstruction results at the image level under large motion scenarios. For DH-FaceReliVid-200, we focused more on the performance of image relighting. Therefore, we selected reference images across videos to test the final results and specifically assessed the accuracy of reconstructed images. For each dataset comparison, we evaluated the metrics of UniAvatar under different modalities of control signals.

\begin{figure}[t]  
    \centering
    \includegraphics[width=\linewidth]{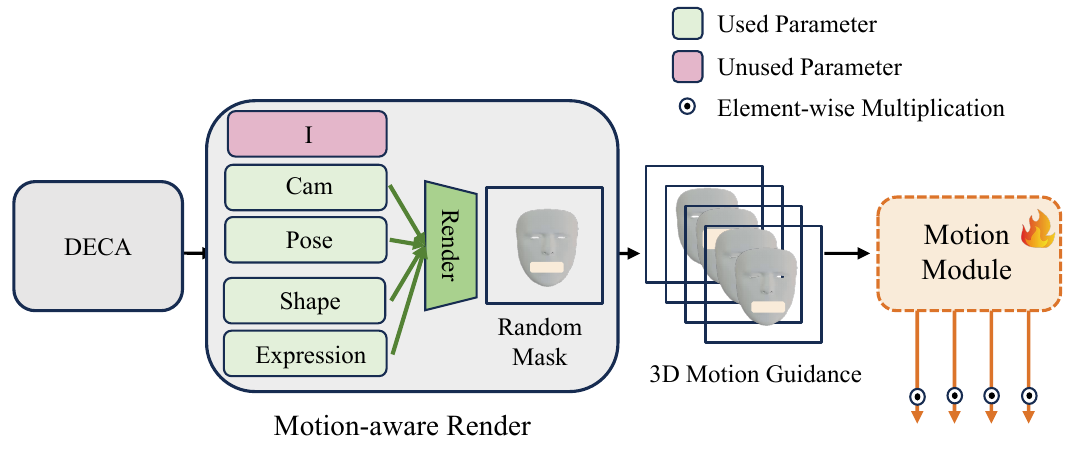}  
    \caption{3D motion guidance is rendered without illumination information to ensure that this module focuses solely on extracting motion information.}
    \label{fig:motion-guidance}
    \vspace{-5mm}
\end{figure}
\begin{figure}[t]  
    \centering
    \includegraphics[width=\linewidth]{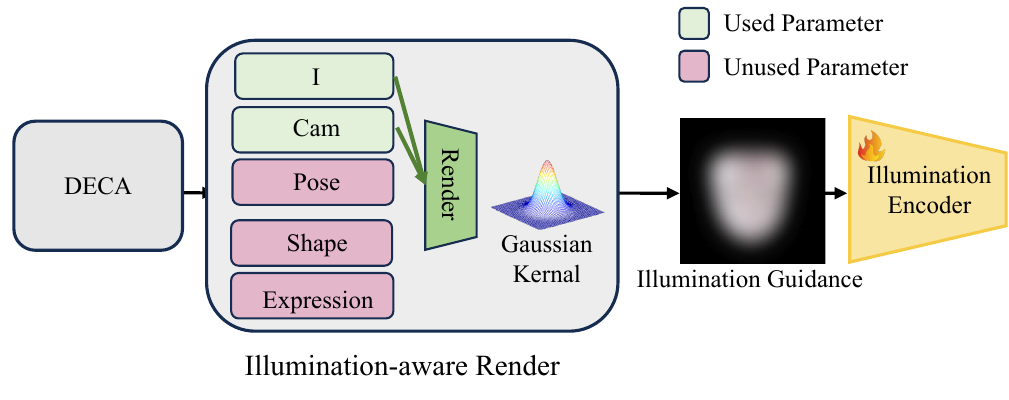}  
    \caption{Given a target video, we render a illumination condition image by extracting the lighting and camera parameters from the first frame, with all other parameters set to initial values to prevent motion information from interfering with the global illumination information.}
    \label{fig:illumination-guidance}
    \vspace{-5mm}
\end{figure}
\subsection{Training and Inference}
\noindent\textbf{Training.} Training was conducted on a computing platform equipped with 8 NVIDIA A800 GPUs. Our approach employs a two-stage training process. In the first stage, the model generates motion video frames based on reference images, illumination conditions, and 3D motion conditions, while keeping the parameters of the VAE encoder/decoder and facial image encoder fixed. For the first 10,000 steps, only the 3D motion conditions are trained, using the HDTF, CelebV-HQ, CelebV-Text, and DH-FaceDrasMvVid-100 datasets, with the lip region of the 3D motion conditions masked at a probability of 0.4. The training focuses on the spatial and cross-attention modules in ReferenceNet, the denoising U-Net, and the 3D motion encoder, where the illumination guidance is initialized as learnable parameters. In the subsequent 5,000 steps, the DH-FaceReliVid-200 dataset is added, and the parameters of the 3D motion encoder are fixed. Throughout this first stage, the batch size remains at 8, and the image resolution is set to be 512x512. In the second stage, each training instance generates 14 video frames. The latent representations from the motion module are concatenated with the first 2 ground truth frames to ensure video continuity. During both training stages, a learning rate of 1e-5 is used. To enhance video generation, the 3D motion conditions are dropped at a probability of 0.2, and the reference images, illumination conditions, guiding audio, and motion frames are dropped at a probability of 0.05 during training.

\noindent\textbf{Inference.} During inference, continuity between sequences is ensured by concatenating the noisy latent period with the feature maps of the last 2 motion frames from the previous step in the motion module.
\section{Network architectures}\label{sec:C}
\subsection{3D Motion Guidance}
The motion module architecture is based on the U-Net network but includes only the downsampling module, consisting of 6 blocks, each immediately followed by a downsampling convolution layer. The first four blocks are composed of residual blocks, and the last two blocks incorporate an attention layer after the residual blocks. We extract shape, pose, expression, camera, and lighting information through DECA \cite{deca} and generate motion conditions consistent with the target video by replacing all other information except lighting. The motion conditions input to the motion module are normalized to the range [-1, 1] and have a tensor shape of 3×512×512. After being extracted by each block in the motion module, a motion representation is output before input to the downsampling layer, with the tensor shape of the motion representation output by every two blocks being consistent. The tensor shapes output by every two blocks are 4096×320, 1024×640, and 256×1280. The motion representations output by each block are element-wise multiplied with the outputs of each spatial attention layer (a total of 6) of the denoising U-Net for motion condition fusion.

\subsection{Illumination Guidance}

The input to the lighting module is also extracted from DECA to obtain shape, pose, expression, camera, and lighting information, and only the lighting information is replaced to generate preliminary lighting conditions consistent with the target lighting. Subsequently, a Gaussian blur is added to the preliminary lighting conditions.
\begin{equation}
G_{i,j} = -\frac{1}{2\pi \sigma^2}  exp  (- \frac{(i-n)^2 + (j-n)^2}{2\sigma^2})
\end{equation}
\begin{equation}
P^{I} = P^{I}_{0} \odot G
\end{equation}

$G_{i,j}$ is the value at pixel position $(i, j)$ in the Gaussian kernel, where $n$ is the center of the Gaussian kernel. $\sigma$ is the noise variance, typically set to one-third of the Gaussian kernel radius, which is 15 in this case. $P^{I}$is the image after Gaussian blur processing.
\begin{figure}[t]  
    \centering
    \includegraphics[width=\linewidth]{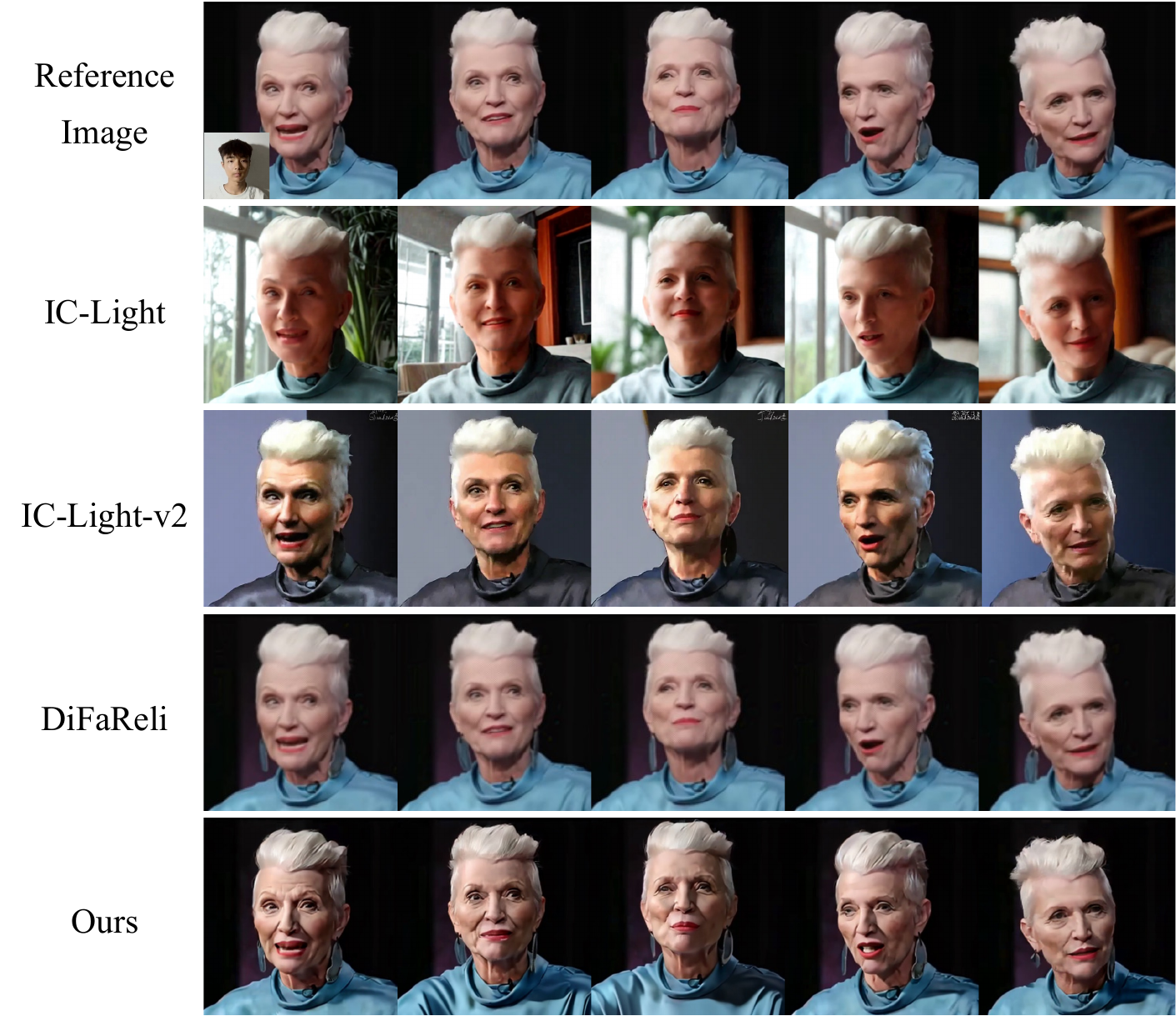}  
    \caption{Compared to other relighting methods, UniAvatar can provide more accurate lighting effects while maintaining a stable background and consistent identity.}
    \label{fig:IC}
    \vspace{-3mm}
\end{figure}
The final illumination conditions input to the illumination module are normalized to the range [-1, 1] and have a tensor shape of 3×512×512. The illumination module consists of convolution layers and an attention module. The illumination conditions undergo illumination information extraction through eight layers of convolution after being input to the convolution layers, followed by an attention layer to more accurately capture illumination information. Finally, the output of the convolution layer initialized to zero results in a illumination representation with a tensor shape of 320×64×64.
\section{Additional Results}\label{sec:D}

To further support the conclusions drawn in the main paper, we provide additional results in this section, showcasing:
\begin{table}[t]
\setlength\tabcolsep{4pt}
\centering
\caption{Ablation experiment on dataset efficiency. "w/o DM" means training without the DH-FaceDrasMvVid-100 dataset, while "w DM" means training with the DH-FaceDrasMvVid-100 dataset.}
\label{Table:dras}
\resizebox{\linewidth}{!}{
\begin{tabular}{cccccc}
\toprule
\multirow{2}{*}{Method} & \multicolumn{5}{c}{DH-FaceDrasMvVid-100}  \\ \cline{2-6} 
                        & FID$\downarrow$    & FVD$\downarrow$     & LPIPS$\downarrow$ & PSNR$\uparrow$   & AKD$\downarrow$   \\ \hline
Ours (w/o DM)           & 49.351 & 413.21  & 0.197 & 20.727  & 3.972 \\
Ours ( w DM)            & 36.644 & 311.491 & 0.122 & 24.477 & 2.459 \\ \toprule
\end{tabular}
}
\vspace{-3mm}
\end{table}
\begin{table}[t]
\setlength\tabcolsep{4pt}
\centering
\caption{Ablation experiment on dataset efficiency. "w/o RL" means training without the DH-FaceReliVid-200 dataset, while "w DM" means training with the DH-FaceReliVid-200 dataset.}
\label{Table:reli}
\resizebox{\linewidth}{!}{
\begin{tabular}{ccccc}
\toprule
\multirow{2}{*}{Method} & \multicolumn{4}{c}{DH-FaceReliVid-200} \\ \cline{2-5} 
                        & FID$\downarrow$      & FVD$\downarrow$       & LPIPS$\downarrow$  & PSNR$\uparrow$   \\ \hline
Ours (w/o RL)           & 114.371  & 1123.631  & 0.301  & 16.541 \\
Ours ( w RL)            & 74.372   & 607.725   & 0.175  & 22.521 \\ \toprule
\end{tabular}
}
\vspace{-3mm}

\end{table}
\begin{enumerate}
    \item Comparison with other relighting methods. UniAvatar can provide more accurate lighting effects and a more stable background while maintaining the person's identity. It is worth noting that, upon testing, DiFaReli \cite{difareli} failed to deliver satisfactory lighting effects, a problem that has also been frequently raised in the Issues section on its project page. (Figure \ref{fig:IC});

    \item Dataset effectiveness analysis was conducted by comparing the quantitative results of training with and without the DH-FaceDrasMvVid-100 and DH-FaceReliVid-200 datasets. Table \ref{Table:dras} shows that training with the DH-FaceDrasMvVid-100 dataset enhances the model's performance in generating results under large motion control signals, while Table \ref{Table:reli} demonstrates that using the DH-FaceReliVid-200 dataset improves the method's performance under varying lighting conditions.;

    \item Initially, to maintain a fixed background during motion and address inconsistencies in the background under different lighting conditions in the dataset, we referred to the channel-wise fusion approach by DiFaReli \cite{difareli}, inputting the motion conditions along with the background image into the motion module. The background image is detected using a face segmentation algorithm. Hair, neck, and clothing all belong to the background part. As shown in Figure \ref{fig:bg}, although the background can be fixed, it also limits the pose movements of the character, allowing only expression control within the facial region of the reference image. Therefore, we removed the input of the background and controlled the background solely through the reference net and the Masked Cross Source Sampling method, which showed that the character could perform flexible pose and expression control through the motion conditions.


\end{enumerate}
\begin{figure}[t]  
    \centering
    \includegraphics[width=\linewidth]{sec/showcase/bg.pdf}  
    \caption{Comparison between different background enhancement methods: our proposed Masked Cross-Source Sampling strategy enables more flexible background preservation under motion control.}
    \label{fig:bg}
    \vspace{-3mm}
\end{figure}



{
    \small
    \bibliographystyle{ieeenat_fullname}
    \bibliography{main}
}